\documentclass[11pt]{article}

\usepackage[preprint]{acl} %

\usepackage{times}
\usepackage{latexsym}

\usepackage[T1]{fontenc}

\usepackage[utf8]{inputenc}

\usepackage{microtype}

\usepackage{inconsolata}

\usepackage{graphicx}

\usepackage{placeins}
\usepackage{enumitem,amsmath}

\title{Does Modern Standard Arabic (MSA) Dominate Arabic Dialects in LLMs?\\
A Representation-Level Analysis}

\author{
    \textbf{Abdu Sallouh}\textsuperscript{\rm 1}, 
    \textbf{Nicholas Popovi\v{c}}\textsuperscript{\rm 2}\thanks{Work conducted while at TU Dresden \& ScaDS.AI}, \and
    \textbf{Michael F\"{a}rber}\textsuperscript{\rm 1,3} \\
    \textsuperscript{\rm 1}Technical University of Dresden \\
    \textsuperscript{\rm 2}Mohamed bin Zayed University of Artificial Intelligence \\
    \textsuperscript{\rm 3}ScaDS.AI, TU Dresden \\
    \texttt{abdu.sallouh@mailbox.tu-dresden.de} \\
    \texttt{nicholas.popovic@mbzuai.ac.ae} \\
    \texttt{michael.faerber@tu-dresden.de}
}

\begin{document}

\maketitle

\begin{abstract}
Large language models (LLMs) often default to Modern Standard Arabic (MSA) when generating Arabic, even when prompted with dialectal Arabic. A natural explanation is that their internal representations are dominated by MSA. We test this hypothesis by adapting the language-dominance framework of \citet{shani2025language} to 26 Arabic varieties.
Across layers and model families, we find no evidence that MSA acts as a dominant internal representation for Arabic dialects. Instead, dialect representations form a dense and highly overlapping space: normalized mutual information drops sharply compared to patterns reported for more distinct languages. Moreover, the strongest separability effects are not confined to intermediate layers, but can shift toward later layers depending on the architecture. These findings challenge a common interpretation of MSA-biased generation: output preference does not necessarily reveal internal representational dominance. Analyses of multilingual and dialectal LLMs should therefore distinguish generation bias from the geometry of internal representations.
\end{abstract}

\begin{figure*}[t]
\centering
\includegraphics[width=\linewidth]{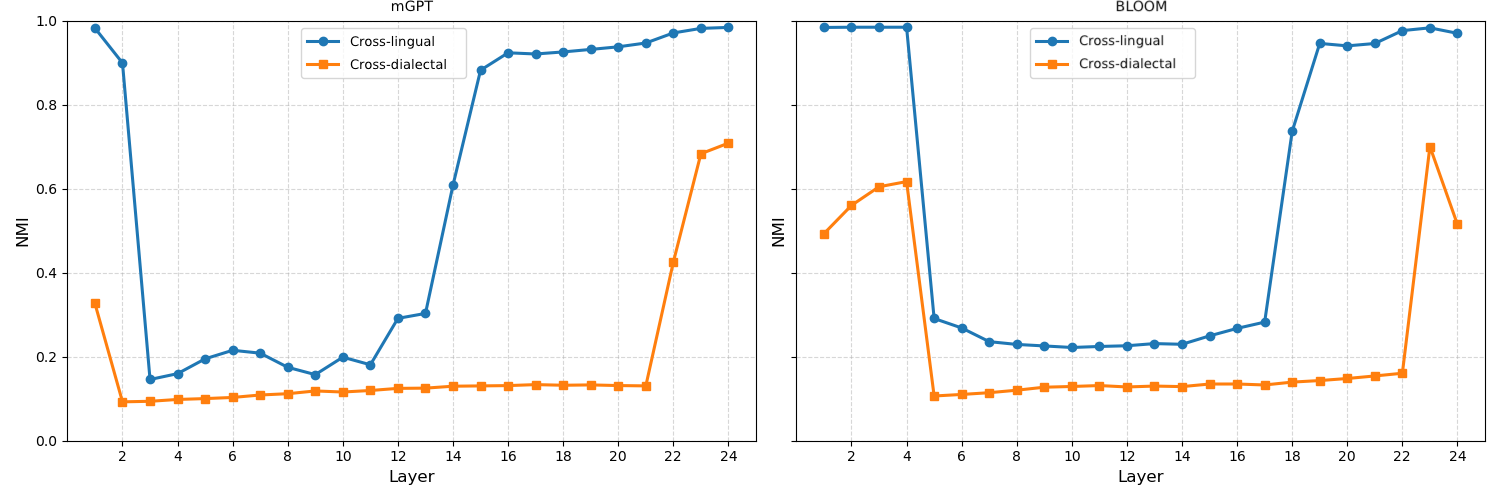}
\caption{Layer-wise NMI values for cross-lingual vs. cross-dialectal representations for mGPT (left) and BLOOM (right). Dialectal NMI is lower and remains low over a wider middle-layer range. Cross-lingual representations were computed using Parallel Universal Dependencies (PUD) treebanks \citep{zeman2017conll} (following \citet{shani2025language}), cross-dialectal representations were computed using the MADAR corpus \citep{bouamor2018madar}.}
\label{fig:nmi}
\end{figure*}

\section{Introduction}

Large language models (LLMs) increasingly support Arabic, including dialectal varieties such as Egyptian, Levantine, Gulf, and Maghrebi Arabic, in applications such as translation, search, and conversational agents. Yet recent work shows a persistent asymmetry: models may understand dialectal input, but often default to Modern Standard Arabic (MSA) when generating output \citep{robinsonetal2025alqasida}. This raises a natural mechanistic question: does MSA dominate only the model's output behavior, or does it also act as a dominant internal representation when the model processes Arabic dialects?

A recent line of work studies such questions through language dominance: the hypothesis that multilingual models internally route cross-lingual processing through a dominant language seen more frequently during pretraining \citep{shani2025language}. Arabic provides a particularly challenging test case for this idea. Unlike clearly separated languages, Arabic varieties form a closely related dialect continuum with shared vocabulary, morphology, and frequent interaction with MSA. If LLMs internally process dialectal Arabic through MSA, this would offer a direct representational explanation for MSA-biased generation. If they do not, then output preference cannot be read as evidence of internal dominance.

We test this hypothesis by adapting language-dominance probing from distinct languages to Arabic varieties. Using a parallel dialectal setup covering 25 Arabic dialects and MSA, we analyze hidden representations across layers and model families to ask whether any variety acts as a dominant internal reference point. Our results show no evidence of MSA or dialectal dominance. Instead, Arabic varieties form a dense and highly overlapping representational space, while layer-wise separability differs across architectures. This suggests that MSA-biased generation and internal representational dominance are distinct phenomena.

Overall, this paper makes three contributions:
\begin{itemize}[leftmargin=*,itemsep=1pt,topsep=1pt,parsep=1pt]
    \item We adapt language-dominance probing to the fine-grained setting of closely related Arabic varieties.
    \item We show that MSA-biased generation does not imply MSA-dominated internal representations.
    \item We identify dense dialectal overlap as a key property of Arabic representations in LLMs, suggesting that generation bias and internal representation geometry must be analyzed separately.\footnote{Our code and data are publicly available at \url{https://github.com/abduABS/MSA_Dominance_LLM}.}
\end{itemize}

\section{Related Work}

\paragraph{Language Dominance in LLMs } 
Recent work has actively investigated how multilingual LLMs internally represent diverse languages. Earlier studies, utilizing logit lens analyses, reported strong indications of language dominance within intermediate model layers, hypothesizing an English-centric representational space \citep{zhang2023donttrustchatgptquestion, wendleretal2024llamas, zhongetal2025language}. However, \citet{belrose2025elicitinglatentpredictionstransformers} demonstrated that logit lens approaches can yield misleading interpretations when applied to complex, modern LLM architectures. To overcome these limitations, \citet{shani2025language} introduced an internal dominance probing framework based on Gaussian Mixture Models (GMMs), suggesting that no single language truly dominates cross-lingual spaces. In this work, we adapt their GMM framework to investigate whether this lack of dominance holds true for closely related linguistic varieties.

\paragraph{Representational Challenges in Arabic Dialects } 
Arabic provides a challenging case study for internal representation due to its diglossic nature. While frequently classified as a high-resource language \citep{nicholasbhatia2023lost}, \citet{alzubaidi2025evaluatingarabiclargelanguage} note that analyses of how LLMs process Arabic internally remain a significant gap. %
Recent studies have begun probing this space, revealing that structural overlap heavily influences model behavior. For instance, \citet{khalak-etal-2026-fusha} observed cross-lingual transfer between MSA and dialects, though disproportionately tied to geographic proximity. Concurrently, \citet{elshabrawy2025alignmenthurtsdecouplingrepresentational} provided causal evidence that excessive representational entanglement with MSA actively restricts generative capacity for Arabic dialects, demonstrating that decoupling dialectal subspaces from MSA during fine-tuning improves dialectal generation quality. Given MSA’s natural prominence in post-training datasets \citep{robinsonetal2025alqasida}, whether this output bias stems from a true representational dominance within the internal layers, or merely dense entanglement, remains an unexplored gap that our layer-wise analysis aims to resolve.

\section{Experimental Framework}
\label{sec:methodology}

We adapt the language-dominance framework of \citet{shani2025language} from a cross-lingual to a cross-dialectal setting. The original framework asks whether hidden representations of one language align with the embedding space of another language across transformer layers. We ask the analogous question for Arabic dialects: when an LLM processes dialectal Arabic, do its hidden states remain dialect-specific, or do they systematically align with MSA or another dialect?

We use the MADAR corpus \citep{bouamor2018madar}, a parallel travel-domain dataset covering 25 Arabic dialects and Modern Standard Arabic (MSA). Parallel data are crucial because they allow us to compare representations of semantically equivalent sentences across varieties. Specifically, MADAR was chosen over other parallel Arabic resources because it distinguishes 25 city-level varieties in addition to MSA, whereas most alternatives cover only a small number of coarse regional groupings. For BLOOM and mGPT, we sample 400 parallel sentences per variety, four times the per-language sample size used by \citet{shani2025language}. Sentences are processed with each model's default tokenizer, and hidden states are extracted across transformer layers.

Our analysis separates two related but distinct questions. First, we measure \textit{separability}: whether Arabic varieties form distinguishable clusters in hidden space. We compute normalized mutual information (NMI) between dialect labels and cluster assignments at each layer; lower NMI indicates that varieties are less separable and more overlapping. Second, we measure \textit{dominance}: whether representations from one variety systematically align with another variety's embedding space. 

Following \citet{shani2025language}, we compute a token-level likelihood ratio $\Lambda(x)$ that compares the posterior probability assigned to the true dialect against the posterior probability assigned to each competing dialect individually. Concretely, for a token $x$
from dialect $d_{true}$ and a competing dialect $c$, we compute $\Lambda(x, c) = p(d_{true} \mid x) / p(c \mid x)$, where the posteriors are derived from the aligned-token setup: since MADAR is parallel, each token $x$ has
a corresponding token at the same position in the semantically equivalent sentence for every other dialect, and we compare the model's posterior over $x$'s own hidden representation against the posteriors over these aligned tokens in competing dialects. This is distinct from dialect classification, as we are not predicting which dialect a sentence belongs to, but rather measuring how strongly a token's representation aligns with its own dialect relative to each competing variety individually. Values of $\Lambda(x, c)$ above one indicate stronger
alignment with the true dialect than with competitor $c$, values below one indicate stronger alignment with $c$, and values near one indicate weak or shared dialect-specific evidence. We do not aggregate $\Lambda$ into a single per-token value. Instead, for each layer we pool every (token, competitor) pair across all tokens and report the fraction of pairs with $\Lambda(x, c) < 1$, i.e., how often a
token's representation aligns more strongly with a competing dialect than with its own.

This distinction is essential for Arabic. Low separability alone does not imply dominance: dialects may overlap because they share script, vocabulary, morphology, and frequent interaction with MSA. We therefore interpret low NMI as evidence of representational overlap, not as evidence that MSA or another dialect dominates the internal representation. A dominance claim requires a directional pattern: representations from one dialect must consistently align more strongly with another variety than with their own.

We additionally report the layer-wise dominance score of \citet{shani2025language}, which measures the average posterior mass a variety attracts from tokens of all other varieties. Let $\mathcal{V}$ denote the set of Arabic varieties under analysis, so that $|\mathcal{V}| = 26$. Under dominance, one variety would show elevated scores across most others while values near the uniform baseline $1/|\mathcal{V}|$ indicate no such pull.

To test whether the observed patterns are architecture-specific, we extend the analysis beyond BLOOM and mGPT to Qwen2.5-7B \citep{qwen2025qwen25technicalreport}, Meta-Llama-3.1-8B-Instruct \citep{dubeyetal2024llama3}, and ArabianGPT-08B-V2 \citep{arabianllm2024arabiangpt}, using 100 parallel samples per variety. This set contrasts multilingual pretraining (Qwen2.5, Llama-3.1) against Arabic-native pretraining (ArabianGPT), rather than aiming to exhaustively benchmark available Arabic LLMs. Replication results, PCA settings, additional model curves, and robustness checks are provided in Appendices~\ref{app:replication}--\ref{app:ablations}.

\begin{figure*}[t]
\centering
\includegraphics[width=\linewidth]{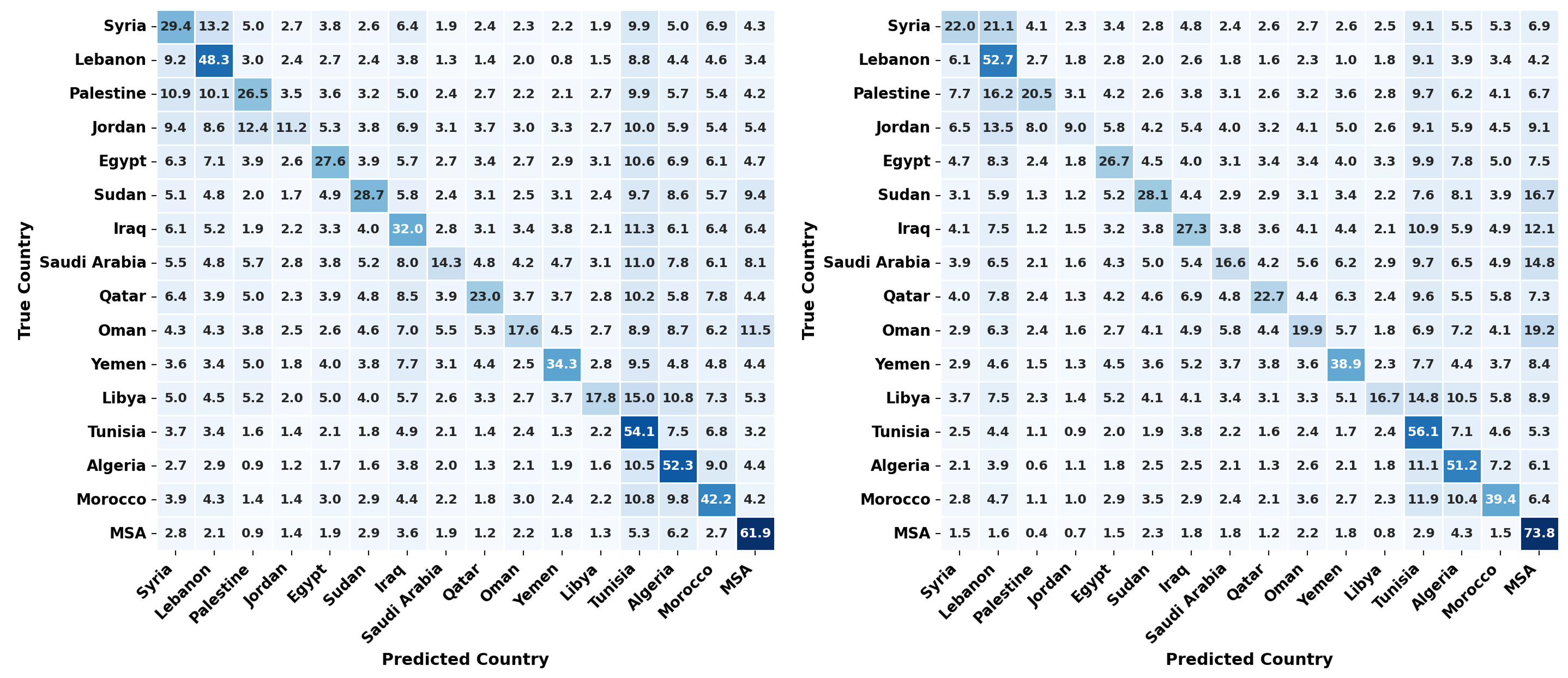}
\caption{Confusion matrices for dialect clustering in mGPT (left) and BLOOM (right) grouped by country. Clear diagonals show that dialect identity is preserved despite low separability.}
\label{fig:confusion}
\end{figure*}

\begin{figure}[t]
\centering
\includegraphics[width=\linewidth]{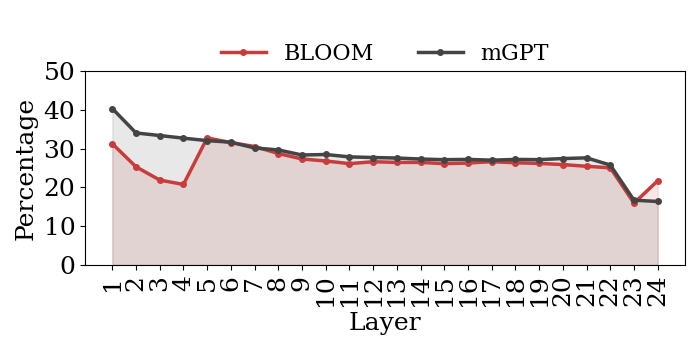}
\caption{Token-level dominance statistics. Percentage of tokens with $\Lambda < 1$.  A mostly flatline percentage of tokens are being assigned to competing dialects. }
\label{fig:lambda}
\end{figure}

\section{Results and Discussion}

\paragraph{Arabic dialects are less separable than distinct languages, but not collapsed.}
Figure~\ref{fig:nmi} shows substantially lower and smoother NMI curves than in our cross-lingual replication of \citet{shani2025language}. Average NMI is around 0.1 for mGPT and 0.2 for BLOOM, and the low-NMI region extends 1--3 layers further at both ends of the middle-layer interval. This indicates a dense representational space in which Arabic dialects are harder to separate than distinct languages. At the same time, the confusion matrices in Figure~\ref{fig:confusion} retain clear diagonals, showing that dialect identity is not erased. A supervised linear probe further confirms this: dialect labels remain predictable above chance despite the low NMI regime (Appendix~\ref{app:ablations}). The strongest off-diagonal overlap appears for geographically close pairs such as Rabat--Fes and Tunis--Sfax. Full confusion matrices and per-layer heatmaps are provided in Appendix~\ref{app:full_figures}.

\paragraph{Dialect overlap reflects more than a failed clustering signal.}
The lower separability is consistent with the high written-form similarity of Arabic dialects. \citet{keleg-etal-2025-revisiting} report that many dialect-specific sentences can be plausibly assigned to multiple dialects, and the parallel MADAR setup further controls semantic variation by design. As a result, the remaining signal is largely carried by surface-form, morphological, and lexical differences. The confusion matrices support this interpretation, but also show that representational overlap is not reducible to a simple lexical-similarity ranking: geographically local pairs stand out strongly, while some dialect groups known to be lexically close remain comparatively well aligned with their own labels \citep{kwaik_lexical_2018}. This suggests that LLM hidden states encode a mixture of lexical, regional, and model-specific factors rather than a clean dialect taxonomy. This pattern is also consistent with the broader linguistic view that the boundary between a "dialect" and a "language" is not a sharp line but a continuum, with Arabic varieties positioned along it much like the languages they are often contrasted against.

\paragraph{Representational overlap is not MSA dominance.}
The dominance statistics show that low separability should not be interpreted as MSA or any other dialect acting as an internal hub.  Layer-wise dominance scores remain close to the uniform baseline of 1/$|\mathcal{V}| \approx 0.038$ across layers (Appendix~\ref{app:full_figures}, Figure~\ref{fig:dialect_dominance_heatmap}). The proportion of tokens with $\Lambda < 1$ also stays nearly flat at around 30\% across layers (Figure~\ref{fig:lambda}), rather than forming the middle-layer spike observed for distinct languages (Appendix~\ref{app:full_figures}, Figure~\ref{fig:rep_pairwise}). Many of these tokens receive very low scores, suggesting broad cross-dialect ambiguity rather than a stable directional pull toward one variety. A low score indicates stronger alignment with a particular competitor, not with a single variety across all pairs. Thus, the models exhibit dialectal overlap without evidence that hidden states are routed through MSA. Note that this test is symmetric rather than MSA-specific: we compute $\Lambda$ for every dialect pair at every layer, without assuming in advance which variety would dominate. The same probe produces the middle-layer spike on distinct languages (Appendix~\ref{app:full_figures}, Figure~\ref{fig:rep_pairwise}), which establishes that it detects directional alignment when the effect is strong enough to appear.

\paragraph{MSA-biased generation is therefore not explained by internal MSA dominance.}
This is the central result. Prior work reports that LLMs often default to MSA when generating Arabic dialects \citep{robinsonetal2025alqasida}, but our analysis does not find a corresponding representational dominance of MSA in hidden states. Under the applied probing framework, MSA behaves as one variety among many rather than as an implicit internal translator. This does not mean that output bias is unrelated to training data or decoding behavior; rather, it shows that output preference should not be read directly as evidence of a dominant internal representational pathway.

\paragraph{The location of dialectal overlap is architecture-dependent.}
The additional models preserve the absence of strong dominance effects, but differ in where dialectal separability changes across depth. Meta-Llama-3.1-8B-Instruct maintains relatively stable dialect separability. Qwen2.5-7B shows an early sharp drop followed by a modest recovery toward the final layer. ArabianGPT-08B-V2 shows the most distinctive pattern: NMI remains high until later layers and then collapses sharply, with a small increase in $\Lambda < 1$ tokens. Since ArabianGPT is trained predominantly on Arabic data, this suggests that Arabic-focused training does not necessarily make dialects uniformly more separable. It may instead shift dialectal compression toward later layers.

\paragraph{Robustness checks and implication.}
Ablations with POS filtering and sentence-prefix removal preserve the main trends (Appendix~\ref{app:ablations}). This suggests that the observed overlap is not driven only by function words or sentence-initial artifacts. Taken together, the results show that generation bias and internal representation geometry must be analyzed separately. For dialectal Arabic NLP, the challenge is to understand how closely related varieties are packed into a shared representational space and how this affects dialect-sensitive generation.

\section{Conclusion}

We examined whether LLMs' tendency to generate Modern Standard Arabic (MSA) for dialectal Arabic reflects internal representational dominance. Adapting the language-dominance framework of \citet{shani2025language} to 25 Arabic dialects and MSA, we find no evidence that MSA, or any dialect, acts as a dominant internal reference point. Instead, Arabic dialects form a dense and highly overlapping representational space: dialectal information remains present, but it is not organized around a single dominant variety.

This changes how MSA-biased generation should be interpreted. Output preference does not necessarily reveal the geometry of hidden representations, and generation bias should not be treated as direct evidence of internal routing. If internal dominance is not the driver, the bias plausibly originates elsewhere in the generation pipeline: possible contributors include the disproportionate share of MSA in pretraining and instruction-tuning data, the comparatively standardized orthography of MSA relative to unstandardized dialectal spelling conventions, or decoding-time preferences shaped by RLHF and higher-frequency token paths, though disentangling these explanations is beyond the scope of this work. More broadly, analyses of multilingual and dialectal LLMs should distinguish what models prefer to generate from how they internally organize closely related linguistic varieties.

\section{Limitations} 
This study faced several limitations. First, the experiments use 400 sentences per dialect for BLOOM and mGPT and 100 sentences per dialect for the additional models from the MADAR corpus \citep{bouamor2018madar} due to performance constraints. While MADAR is parallel and fine-grained across 26 varieties, it is restricted to the travel domain, which may not reflect the full range of dialectal variation in natural text. 

Furthermore, 400 sentences per dialect is a relatively moderate sample, and findings should be interpreted as exploratory rather than definitive claims about Arabic dialect representation in LLMs broadly. While MADAR made a substantial effort to recruit diverse writers for each dialect, the total number of writers per dialect remains relatively small, meaning some results may partly reflect writer-specific idiosyncrasies rather than dialect-level patterns, which could modestly inflate separability estimates, though its precise effect on the likelihood-ratio statistics is harder to predict. Additionally, the rule-based POS tagging used in the ablation experiments may introduce tagging inaccuracies compared to fully trained Arabic NLP systems. Future work could explore larger dialect corpora and evaluate the framework on additional Arabic-focused language models which would integrate an efficient backbone and a more sophisticated POS tagger.

Our model selection was designed to contrast multilingual and Arabic-native pretraining rather than to comprehensively survey available Arabic LLMs; evaluating additional recent Arabic-centric models would further strengthen the empirical coverage of this framework and is a natural direction for future work. Finally, positioning Arabic varieties on the dialect–language continuum more precisely would require computing the same separability curves for closely related language families.

\section{Ethics Statement}

This work uses existing datasets and pretrained models and does not involve human subjects or newly collected user data. Dialect labels are treated as dataset annotations for linguistic analysis, not as fixed social or national identities. The results should not be interpreted as ranking dialects, speakers, or communities.

LLM-based assistants were used only for code-development support and grammatical or stylistic feedback. All code, analyses, interpretations, and conclusions were manually verified and remain the authors' responsibility.

\section*{Acknowledgements}

This research was funded by the Center for Scalable Data Analytics and Artificial Intelligence (ScaDS.AI) and the German Federal Ministry of Research, Technology and Space (BMFTR) through the Software Campus project (01IS23070).

\bibliography{custom}

\appendix

\begin{figure}[!t]
    \centering
    \includegraphics[width=1.0\columnwidth]{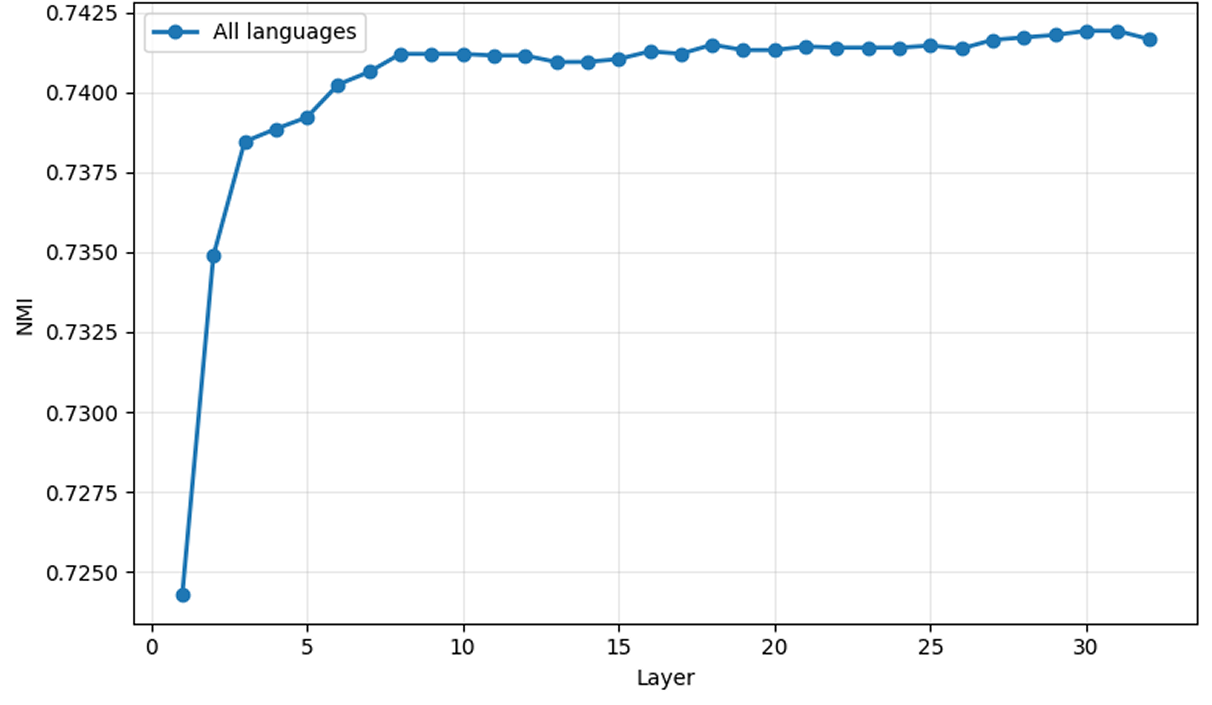}
    \caption{Layer-wise NMI values for Meta-Llama-3.1-8B-Instruct. The model maintains relatively stable dialect separability across middle and later layers.}
    \label{fig:nmi_llama}
    
    \vspace{1.5em}
    
    \includegraphics[width=1.0\columnwidth]{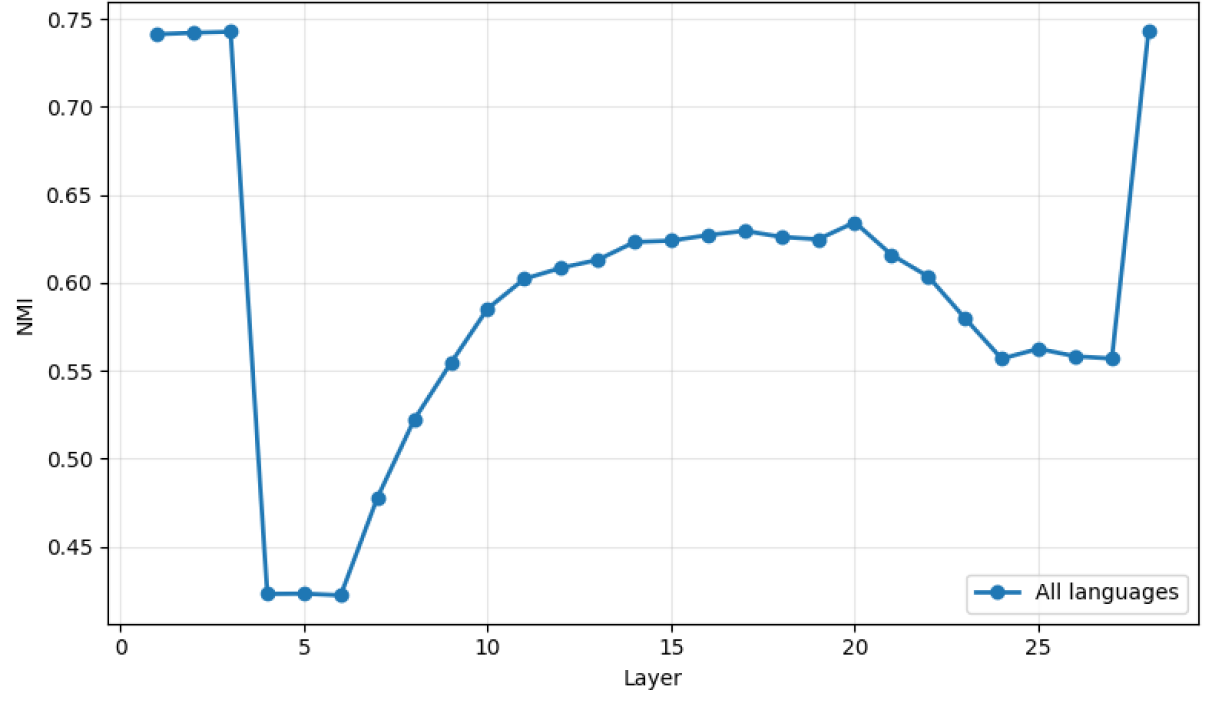}
    \caption{Layer-wise NMI values for Qwen2.5-7B. NMI drops sharply in early layers and partially recovers later.}
    \label{fig:nmi_qwen}
    
    \vspace{1.5em} 
    
    \includegraphics[width=1.0\columnwidth]{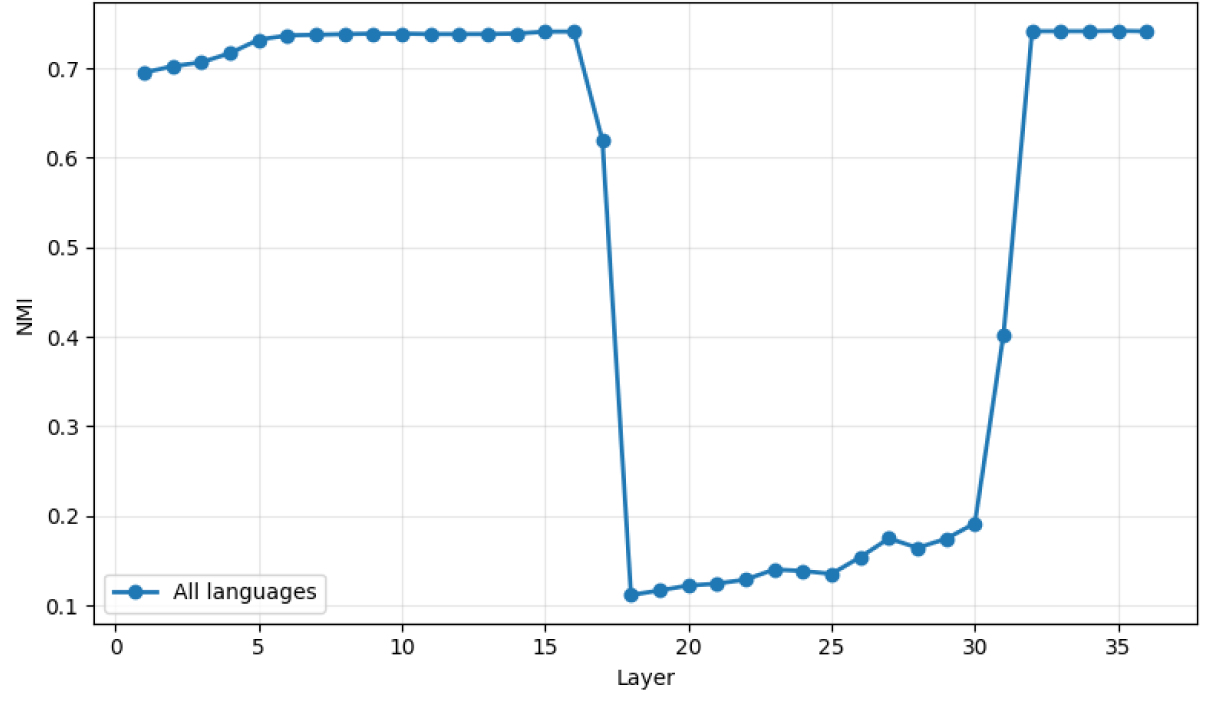}
    \caption{Layer-wise NMI values for ArabianGPT-08B-V2. Dialect separability remains high until later layers before collapsing sharply.}
    \label{fig:nmi_arabiangpt}
    
    \vspace{-1.2em} %
\end{figure}

\section{Replication Setup and Results}
\label{app:replication}

We replicate the experimental setting of \citet{shani2025language} using the \textit{Parallel Universal Dependencies} (PUD) treebanks \citep{zeman2017conll}. Following the original work, we use the multilingual models mGPT \citep{shliazhko2022mgpt} and BLOOM \citep{scao2023bloom} and apply the analysis to the same language set. This replication verifies that our implementation recovers the main trends of the original language-dominance study before extending the framework to Arabic dialects.

To improve numerical stability, we enforce a minimum PCA dimensionality when the retained variance would otherwise collapse the representations to very few components. This modification does not change the logic of the original framework, but prevents unstable mixture modeling in layers with highly anisotropic representations.

The replication aligns closely with the original results. For mGPT, the NMI drop appears between layers 3 and 14, matching the reported middle-layer collapse, although the tail layers are modestly higher in our run by roughly 0.1--0.3 NMI points. For BLOOM, the NMI curves are almost identical to the original, with only small differences at the tails. The replicated confusion matrices and dominance scores also recover the main trends, although we observe minor differences in which languages show weak middle-layer dominance. Token-level $\Lambda$ statistics are broadly consistent across layers and models, with mGPT producing a larger share of very low $\Lambda$ values than BLOOM. Replication figures are provided in Appendix~\ref{app:full_figures}.

\section{Additional NMI Results}
\label{app:additional_nmi}

In addition to BLOOM and mGPT, we evaluate the NMI-based separability analysis on three further LLMs: Meta-Llama-3.1-8B-Instruct, Qwen2.5-7B, and ArabianGPT-08B-V2. The resulting NMI curves are shown in Figures~\ref{fig:nmi_llama}, \ref{fig:nmi_qwen}, and~\ref{fig:nmi_arabiangpt}. While the absence of strong dominance effects remains consistent across models, the layer-wise separability patterns vary substantially across architectures. Some models maintain relatively stable separability, whereas others show sharper collapses in early or late layers.

\section{Ablations and Validation Experiments}
\label{app:ablations}

We conduct additional experiments to test whether the main patterns are artifacts of token type, sentence position, or parameter choices. These analyses support the main-paper conclusion that Arabic dialectal representations are overlapping without showing stable MSA or dialectal dominance.

\subsection{POS Filtering}

Function words such as prepositions, conjunctions, particles, and pronouns are frequently shared across Arabic dialects. If the observed overlap were driven primarily by these high-frequency tokens, removing them should substantially change the dominance statistics. To test this, we use a rule-based POS tagger based on Arabic morphological patterns and lexicon lookup. The tagger identifies ten POS categories: NOUN, VERB, PART, PREP, CONJ, PRON, ADV, NUM, PUNCT, and INTERJ.

For the filtering ablation, we remove PREP, CONJ, PART, PRON, and PUNCT tokens. This removes 64,071 tokens, corresponding to approximately 21.5\% of the corpus, and leaves primarily content-bearing categories such as nouns, verbs, adjectives, adverbs, and numbers.

Applying the dominance analysis to the filtered corpus produces patterns consistent with the main experiments. The absence of stable dialectal or MSA dominance remains unchanged, indicating that the observed overlap is not driven primarily by function words.

\subsection{Sentence-Prefix Removal}

We also test whether sentence-position artifacts influence the dominance statistics. Sentence-initial tokens may reflect repeated openings or common syntactic templates rather than dialect-specific processing. To control for this, we remove the first 20\%, 40\%, and 60\% of tokens from each sentence and recompute the dominance statistics.

Across all settings, the main trends remain stable. Removing larger prefixes slightly reduces the proportion of $\Lambda < 1$ tokens in intermediate layers, suggesting that sentence-initial tokens contribute modestly to cross-dialect mixing. However, the overall absence of stable MSA or dialectal dominance remains unchanged.

\begin{figure}[t]
    \centering
    \includegraphics[width=\columnwidth]{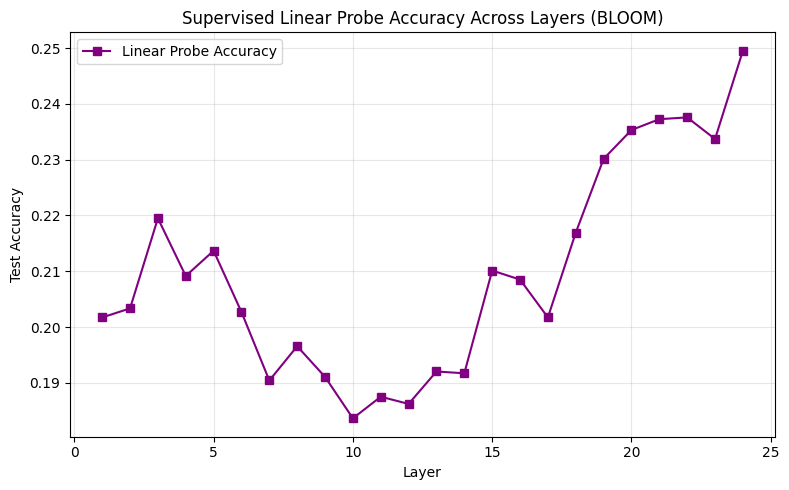}
    \caption{Supervised linear probe accuracy across layers for BLOOM. Accuracy ranges from 0.19 to 0.25.}
    \label{fig:probe_bloom}
\end{figure}

\begin{figure}[t]
    \centering
    \includegraphics[width=\columnwidth]{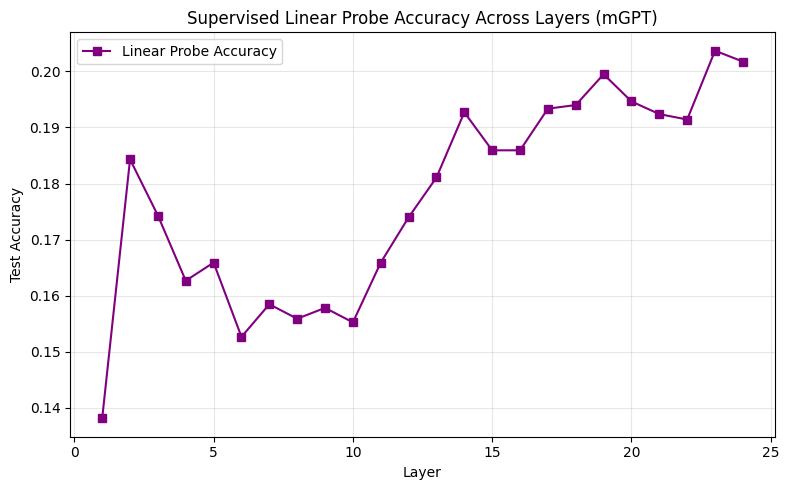}
    \caption{Supervised linear probe accuracy across layers for mGPT. Accuracy ranges from 0.14 to 0.21.}
    \label{fig:probe_mgpt}
\end{figure}

\subsection{PCA Sensitivity Analysis}

We test the robustness of the results to sample size and PCA dimensionality. For mGPT and BLOOM, we vary the sample size from 100 to 400 sentences per dialect in increments of 100. We also vary the minimum PCA dimensionality across 0, 50, 100, 150, 200, and 300 components, where 0 denotes no enforced minimum dimensionality.

The main conclusion remains stable across these configurations. NMI values generally increase with higher principal component counts, suggesting that additional components recover more dialectal structure. However, the absence of stable dominance persists across the tested settings.

\subsection{Supervised Linear Probe}

To verify that dialect information is present even when unsupervised clustering shows low separability, we train a supervised linear probe on hidden states from each mGPT and BLOOM layer. We use logistic regression with z-score standardization, training on 80\% of token-level representations and evaluating on the remaining 20\%.

With 26 variety classes, random chance accuracy is 3.84\%. The probe achieves test accuracies between 0.15 and 0.25 across layers for both models (Figures~\ref{fig:probe_bloom} and~\ref{fig:probe_mgpt}), corresponding to roughly 4--6.5 times random chance. This confirms that dialect information is encoded in representations, even when it is difficult to separate cleanly.

The layer-wise probe accuracy mirrors the NMI results: accuracy is higher in early and late layers, with a dip in the middle layers where NMI collapse is strongest. This convergence between unsupervised and supervised diagnostics supports the interpretation that Arabic dialects are represented in an entangled space rather than being routed through a single dominant variety.

\section{PCA Dimensionality and Experimental Parameters}
\label{app:pca_parameters}

During replication, we observe that the dimensionality-reduction step is sensitive to the variance threshold and model architecture. In all experiments, PCA projections retain 98\% of explained variance. We additionally enforce a minimum dimensionality threshold to prevent the retained space from collapsing to extremely few components in highly anisotropic layers.

Token representations are obtained using the HuggingFace tokenizer with offset mappings enabled to support subword averaging. The tokenization procedure follows the configuration below:

\begin{verbatim}
encoding = tokenizer(
    text,
    return_tensors="pt",
    return_offsets_mapping=True,
    add_special_tokens=True,
    truncation=False
)
\end{verbatim}

The language-space construction procedure uses the following parameters:

\begin{verbatim}
build_language_space(
    layer_idx,
    vectors_dict,
    pca_variance=0.98,
    use_min_pca_dim=True,
    pca_min_dim=X,  # model-specific
    reg_covar=1e-6,
    seed=0
)
\end{verbatim}

The value of \texttt{pca\_min\_dim} is adjusted by model and experiment. For replication on PUD, the minimum PCA dimensionality is 5 for mGPT and 4 for BLOOM. For dialectal analysis, we use 200 components for both models for 400 samples and 100 components for 100 samples. The higher threshold for dialects reflects the greater representational similarity of Arabic dialects compared to distinct languages: more components are needed to retain enough structure for stable mixture modeling and interpretable confusion matrices.

\section{Supplementary Figures}
\label{app:full_figures}

This section provides supplementary figures for replication and dialectal analysis.

\begin{figure*}[t]
\centering
\includegraphics[width=\linewidth]{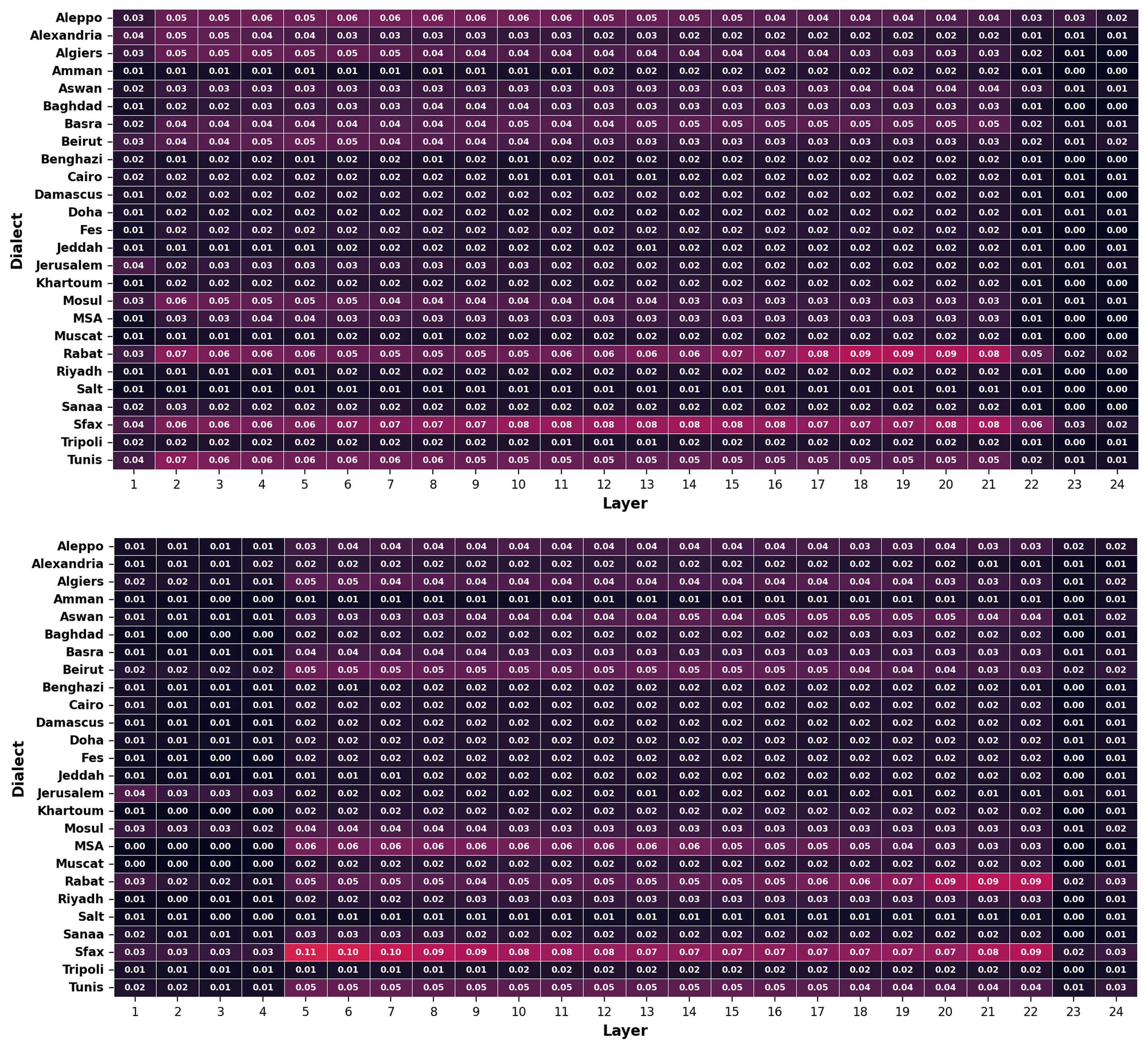}
\caption{Dialect dominance scores across layers for mGPT (top) and BLOOM (bottom). Dominance remains weak across layers.}
\label{fig:dialect_dominance_heatmap}
\end{figure*}

\begin{figure*}[t]
\centering
\includegraphics[width=0.8\linewidth]{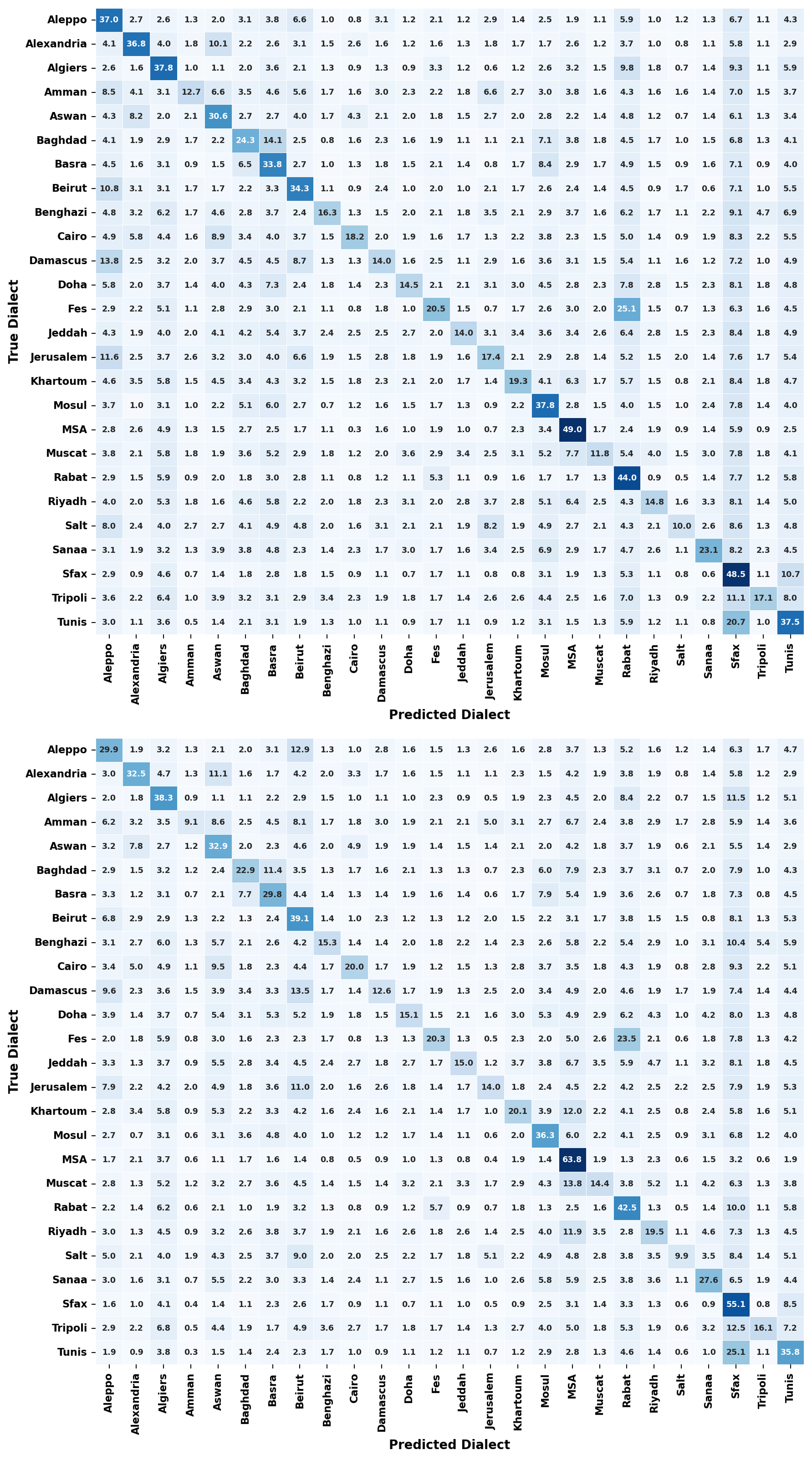}
\caption{Complete dialectal confusion matrices for mGPT (top) and BLOOM (bottom).}
\label{fig:dialect_confusion_matrix}
\end{figure*}

\begin{figure*}[t]
\centering
\includegraphics[width=1.0\linewidth]{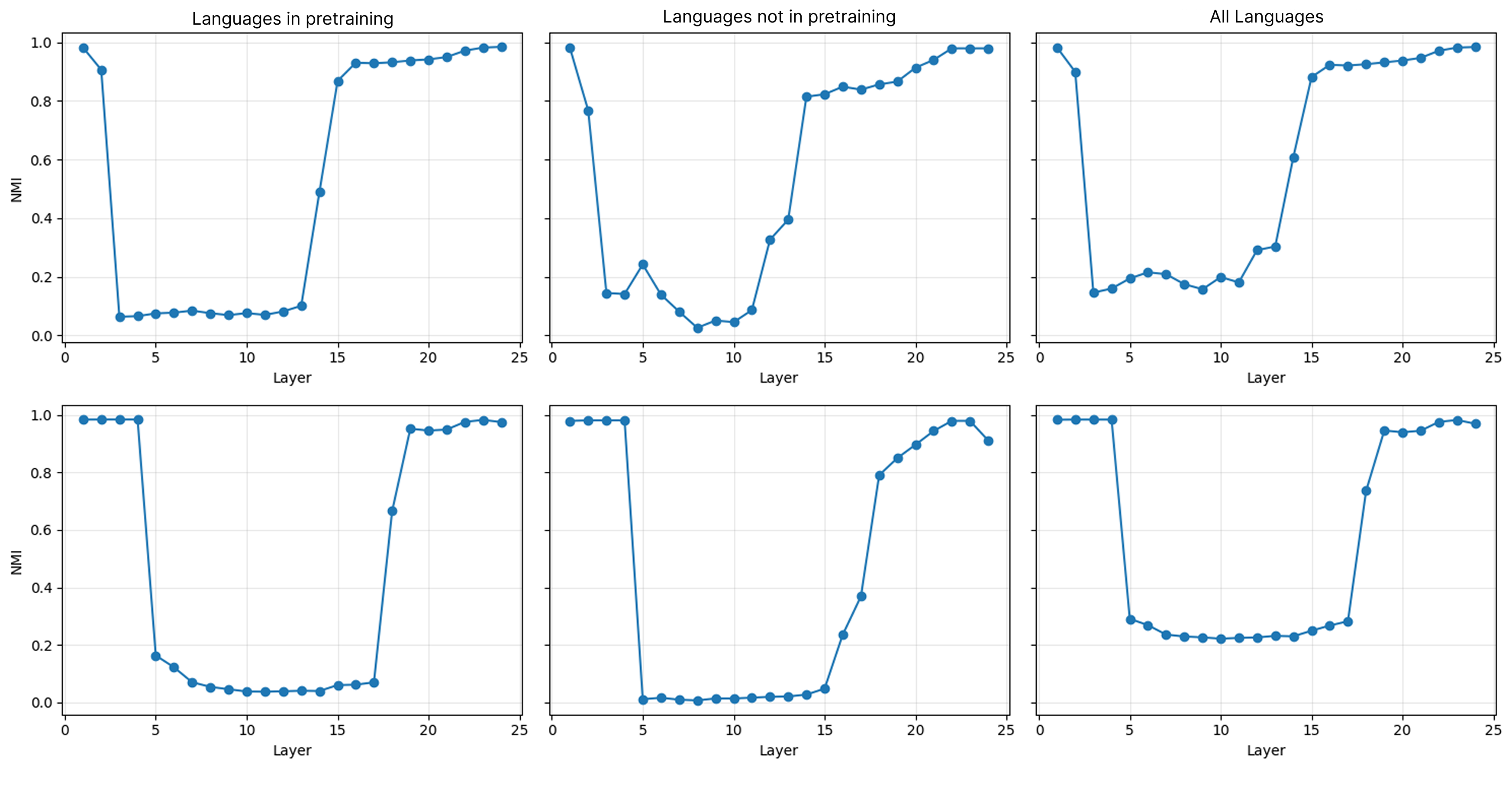}
\caption{Replicated NMI curves for mGPT (top) and BLOOM (bottom).}
\label{fig:rep_nmi}
\end{figure*}

\begin{figure*}[t]
\centering
\includegraphics[width=1.0\linewidth]{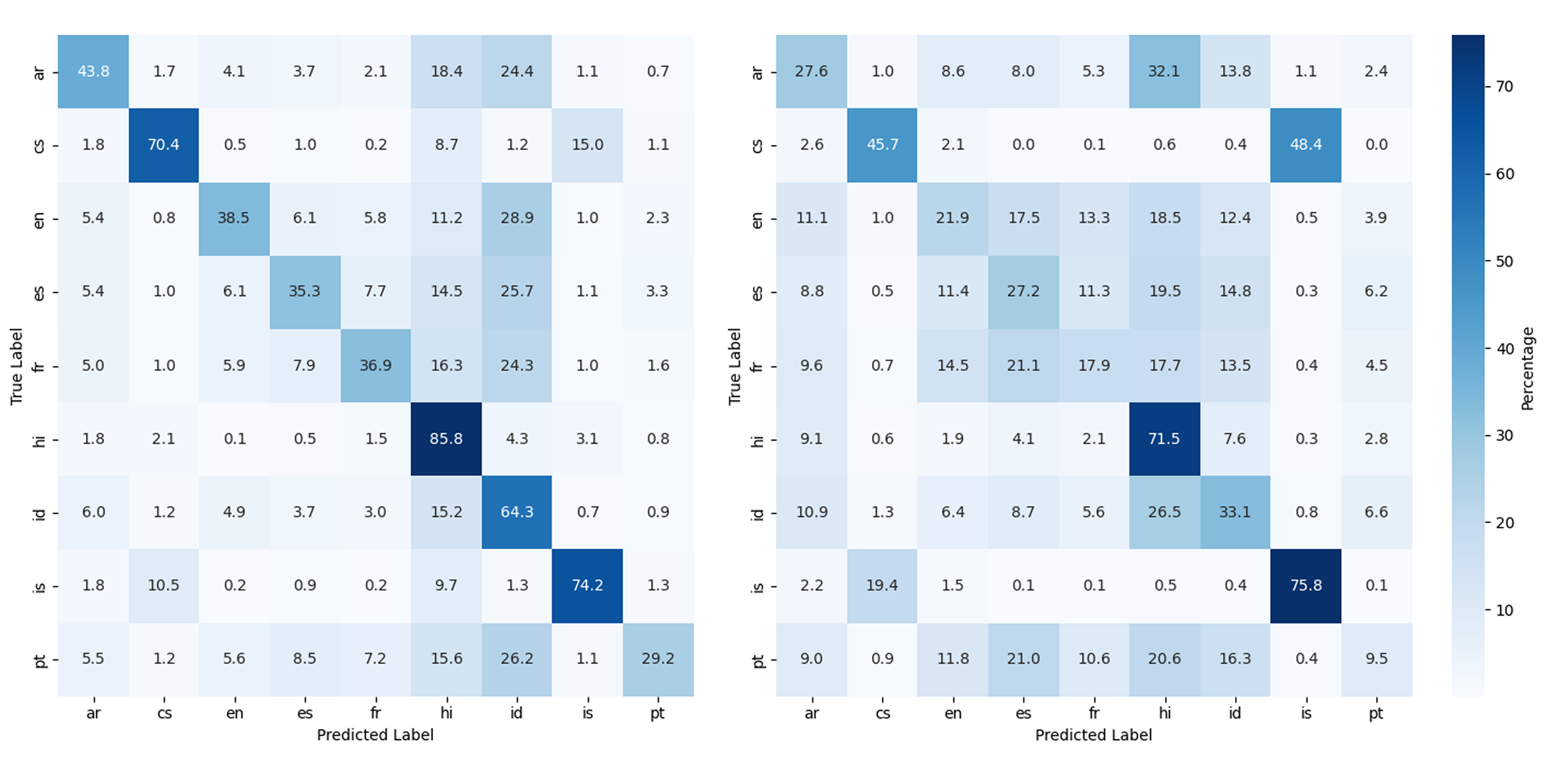}
\caption{Replicated confusion matrices for mGPT (left) and BLOOM (right).}
\label{fig:rep_confusion_matrix}
\end{figure*}

\begin{figure*}[t]
\centering
\includegraphics[width=1.0\linewidth]{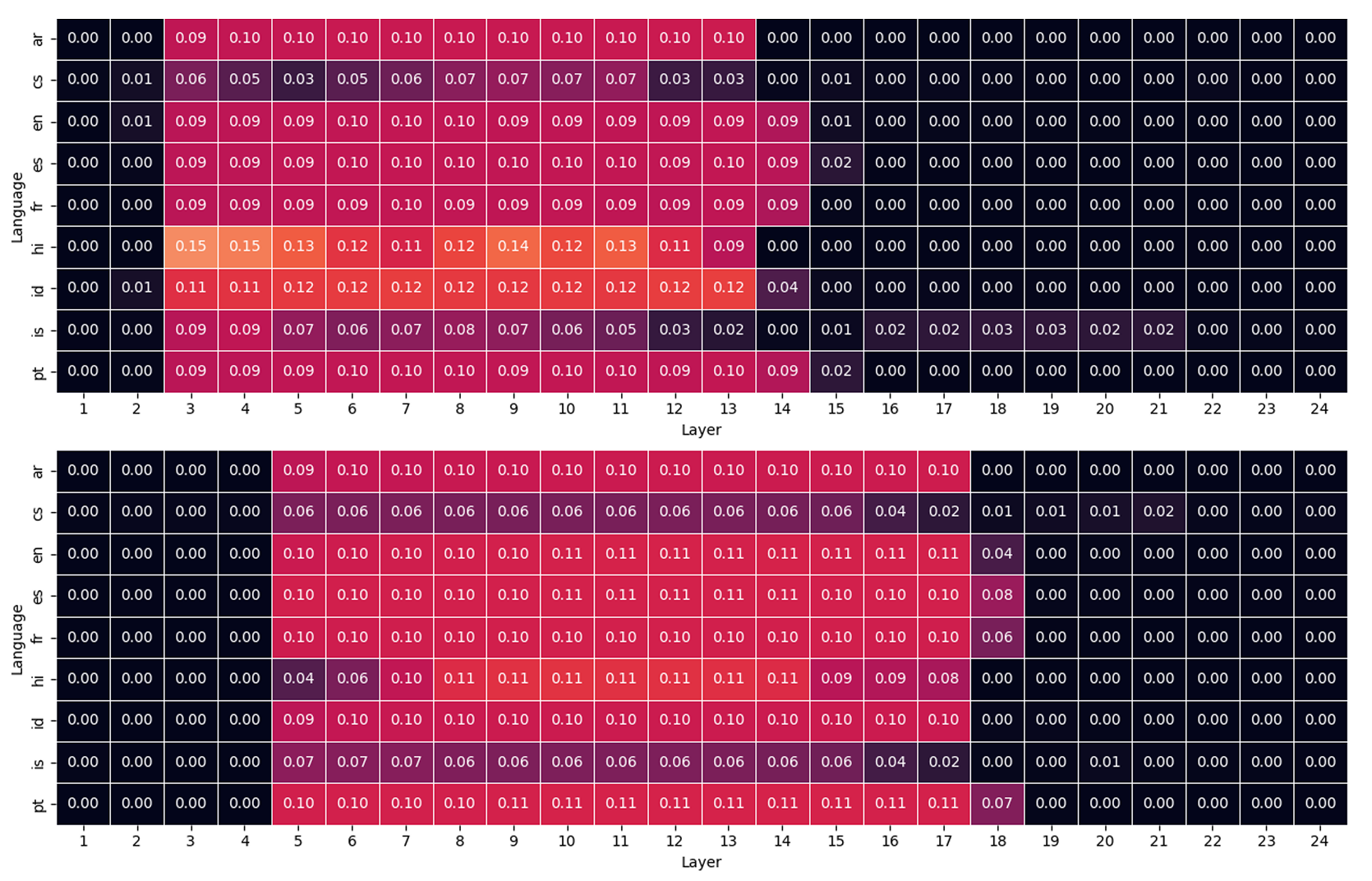}
\caption{Replicated layer-wise dominance scores for mGPT (top) and BLOOM (bottom).}
\label{fig:rep_dominance_heatmap}
\end{figure*}

\begin{figure*}[t]
\centering
\includegraphics[width=1.0\linewidth]{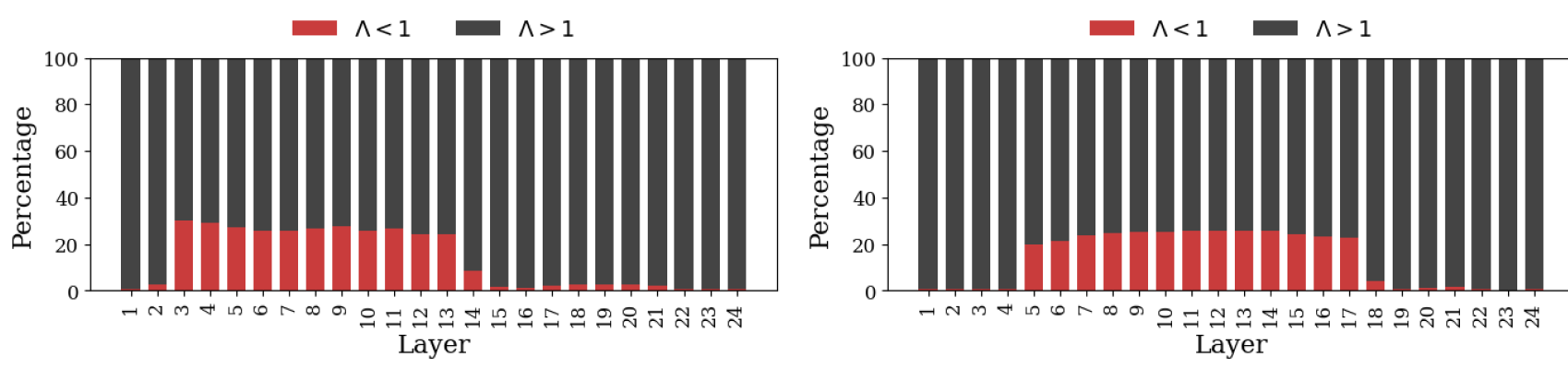}
\caption{Replicated token-level dominance statistics for mGPT (left) and BLOOM (right).}
\label{fig:rep_pairwise}
\end{figure*}

\begin{figure*}[t]
\centering
\includegraphics[width=1.0\linewidth]{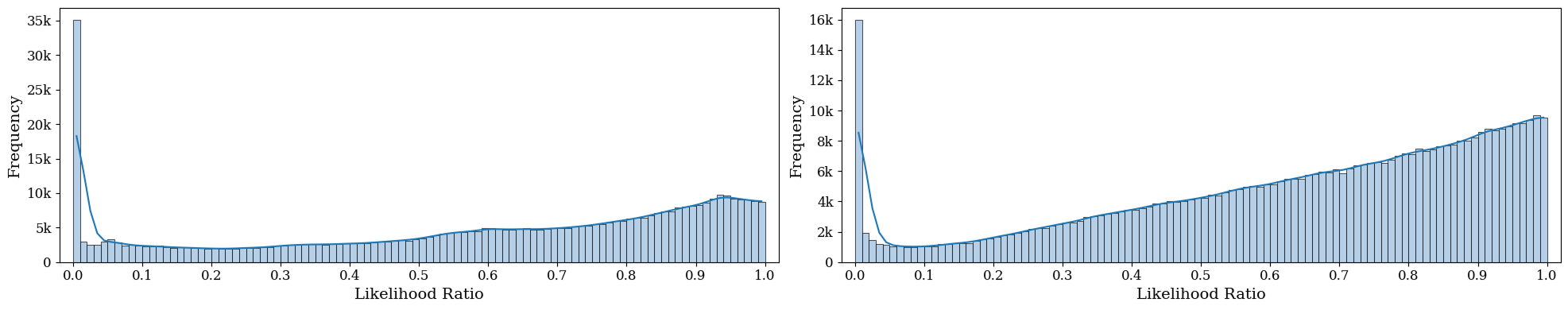}
\caption{Replicated token-level $\Lambda$ histograms for mGPT (left) and BLOOM (right).}
\label{fig:rep_histogram}
\end{figure*}

\end{document}